\documentclass[5p,times,twocolumn]{elsarticle}
\usepackage{lineno}
\usepackage{amsthm,amsmath}
\usepackage{amssymb}
\usepackage{graphicx}
\usepackage{calc}
\usepackage{amstext}
\usepackage{multicol}
\usepackage{pslatex}
\usepackage{textcomp}
\usepackage{multirow}
\usepackage{url}
\usepackage{float}
\usepackage{upquote}
\usepackage{color}
\usepackage{hhline}
\usepackage{mathtools}
\usepackage{array}
\usepackage{tabulary} 
\usepackage{times}
\usepackage{booktabs}
\usepackage{tabularx}
\modulolinenumbers[5]

\journal{Arxiv}

\DeclareUnicodeCharacter{2212}{-}
\begin{document}

\begin{frontmatter}

\title{Pseudo-label Refinement for Improving Self-Supervised Learning Systems}


\author[mymainaddress]{Zia-ur-Rehman}
\ead{mscs20029@itu.edu.pk}
\author[mymainaddress]{Arif Mahmood$^*$}
\ead{arif.mahmood@itu.edu.pk}
\cortext[mycorrespondingauthor]{Corresponding Author}
\author[mysecondaryaddress]{Wenxiong Kang}
\ead{auwxkang@scut.edu.cn}
\address[mymainaddress]{ Center for Artificial Intelligence \& Robot Vision (CAI\&RV), \\
Department of Computer Science, Information Technology University (ITU), Lahore, Pakistan}
\address[mysecondaryaddress]{School of Automation Science and Engineering at South China University of Technology (SCUT)}

\begin{abstract}
Self-supervised learning systems have gained significant attention in recent years by leveraging clustering-based pseudo-labels to provide supervision without the need for human annotations. However, the noise in these pseudo-labels caused by the clustering methods poses a challenge to the learning process leading to degraded performance. In this work, we propose a pseudo-label refinement (SLR) algorithm to address this issue. The cluster labels from the previous epoch are projected to the current epoch cluster-labels space and a linear combination of the new label and the projected label is computed as a soft refined label containing the information from the previous epoch clusters as well as from the current epoch. 
In contrast to the common practice of using the maximum value as a cluster/class indicator, we employ hierarchical clustering on these soft pseudo-labels to generate refined hard-labels. This approach better utilizes the information embedded in the soft labels, outperforming the simple maximum value approach for hard label generation. The effectiveness of the proposed SLR algorithm is evaluated in the context of person re-identification (Re-ID) using unsupervised domain adaptation (UDA). Experimental results demonstrate that the modified Re-ID baseline, incorporating the SLR algorithm, achieves significantly improved mean Average Precision (mAP) performance in various UDA tasks, including real-to-synthetic, synthetic-to-real, and different real-to-real scenarios. These findings highlight the efficacy of the SLR algorithm in enhancing the performance of self-supervised learning systems. 
\end{abstract}

\end{frontmatter}

\section{Introduction}
\label{intro}

Self-supervised learning systems \cite{arazo2020pseudo, 8894380, fan2018unsupervised}, particularly those based on pseudo labels, have gained significant importance in the field of machine learning. These systems offer a powerful approach to training models without the need for extensive labeled datasets, which can be expensive and time-consuming to create. By leveraging unlabeled data, self-supervision-based learning systems enable the model to learn meaningful representations and extract valuable features from the data itself. Pseudo labels, generated through clustering or other techniques, play a crucial role in guiding the learning process by providing supervisory signals. Such approaches have numerous applications across various domains, including computer vision \cite{xu2019self, ramesh2023dissecting}, natural language processing \cite{lan2019albert, fang2020cert,elnaggar2021prottrans}, and unsupervised domain adaptation \cite{tang2021unsupervised, chen2021enhancing, guo2022jac}. It enables tasks such as image classification \cite{azizi2021big, chen2021self}, object detection \cite{lee2019multi,mustikovela2021self, amrani2020self}, object re-identification \cite{wu2020tracklet, khorramshahi2023robust}, and semantic segmentation \cite{wang2020self, hoyer2021three, wang2019self}, where labeled data may be scarce or unavailable. The significance of self-supervised learning lies in its ability to leverage large amounts of unlabeled data, paving the way for more scalable and efficient learning algorithms. As research and advancements in this area continue, the potential for real-world applications and breakthroughs in artificial intelligence grows exponentially.

\begin{figure}[t]
  \centering
  \includegraphics[trim=3cm 0cm 2cm 0cm, clip, width=0.5 \textwidth]{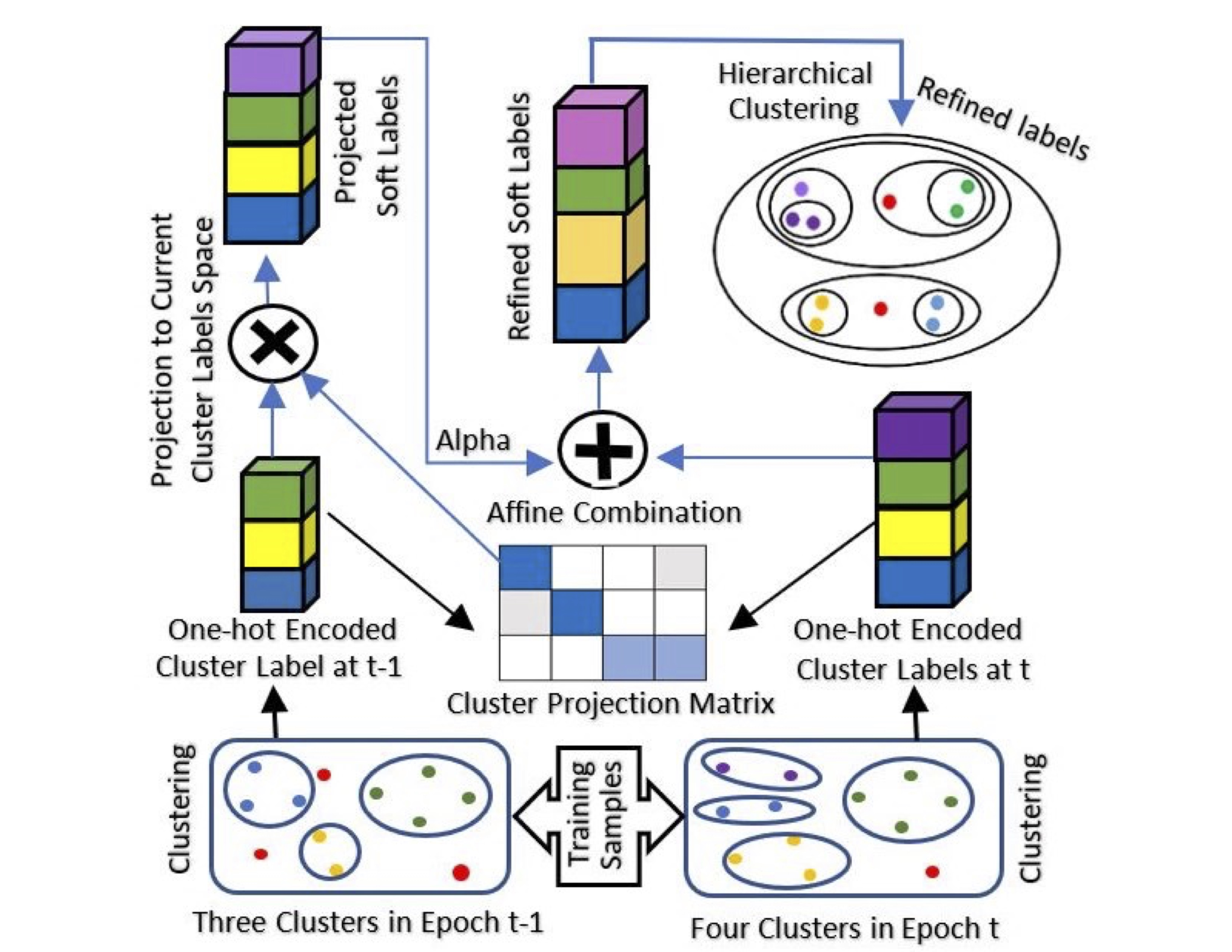}
  \caption[Our Methodology]{An example of the proposed Pseudo Label Refinement (SLR) algorithm:
   in the previous epoch, three clusters were generated while in the current epoch, four clusters are obtained.  Cluster labels are represented using the one-hot encoding scheme. The previous label of an instance is projected to the space of the current labels using a projection matrix. A linear combination of the projected label and the current label is used to get refined soft labels, which are then hierarchically clustered to get refined pseudo labels. }
  \label{fig:proposed}
\end{figure}

Pseudo labels-based self-supervision faces several challenges that need to be addressed for effective and reliable learning systems. One of the key challenges is the generation of accurate and reliable pseudo-labels. Pseudo-labels are derived from clustering or other unsupervised techniques, and they inherently contain noise and errors. The quality of pseudo-labels greatly influences the learning process, and inaccurate labels can lead to degraded performance and misguidance for the model.
Another challenge is the issue of label propagation and error accumulation. Pseudo labels are often used to supervise the learning process iteratively, where the model is trained on one type of labeled data and then used to generate pseudo labels for a different type of unlabeled data. However, errors in the initial pseudo labels can propagate and accumulate throughout the iterations, leading to a loss in performance and potential model divergence. Additionally, pseudo-labels-based self-supervision can be sensitive to the selection of appropriate clustering algorithms. The choice of clustering methods, distance metrics, and thresholds can significantly impact the quality and consistency of pseudo labels.

Addressing these challenges requires research and development of novel techniques to improve the reliability and accuracy of pseudo labels. Methods such as label refinement, cluster ensemble, and active learning have been explored to mitigate label noise and error accumulation. To this end, in the current work, we propose a pseudo-label refinement algorithm (Fig. \ref{fig:proposed}) which can be used to improve the performance of self-supervised learning systems. A projection matrix is learned from the current and the previous epoch clusters which is then used to project the previous cluster labels to the current label space. A linear combination of these projected labels and the current cluster labels is computed as refined soft labels, which are then hierarchically clustered to get refined pseudo labels.

Thus, we improve cluster consistency across consecutive training iterations using this cluster label projection scheme. The improved clustering results in enhanced soft labels, which are then again clustered using a hierarchical approach to get hard pseudo-labels. We observe that using clustering on the soft labels better utilizes the information compared to just using a maximum value for generating hard labels. This step also contributes to performance improvement in a self-supervised learning framework. The refined hard labels are then used for supervision of the learning process. The main contributions of the current work include:
\begin{enumerate}
\item A novel pseudo-label refinement algorithm is proposed to address the issue of pseudo-label noise in self-supervised learning systems employing label consistency over consecutive epochs. Pseudo-labels in the previous epoch are projected to the current epoch cluster-label space. A linear combination of the current label and the projected label is computed as refined soft labels resulting in improved performance.  

\item Soft labels have often been converted to hard labels by just thresholding the maximum value to 1.00 and all other values to 0.00. In contrast, we propose a  hierarchical clustering scheme to convert our refined soft labels to hard labels with further performance improvement. 
    
\item The proposed algorithm is evaluated on the object Re-ID application using unsupervised domain adaptation.  In a wide range of experiments, improved performance is observed compared to the baseline approach. 
\end{enumerate}

The proposed SLR algorithm (Fig. \ref{fig:proposed}) is evaluated in the context of object Re-ID using Unsupervised Domain Adaptation (UDA). Many State-Of-The-Art (SOTA) unsupervised object re-identification methods have utilized clustering-based pseudo-label generation \cite{lin2019bottom, zeng2020hierarchical, yang2021joint, zhai2020ad, chen2021enhancing}. However, in the UDA setting, pseudo-label supervision remains challenging due to increased noise caused by the domain gap between the source and target domains. Additionally, limitations of specific clustering methods can further contribute to the noise in pseudo labels. Moreover, the adaptation of learned features from a smaller source dataset to a larger target dataset with unseen identities becomes limited. Refining the pseudo labels through the proposed SLR algorithm aims to enhance UDA performance. The experiments conducted on three datasets, namely Market1501 \cite{zheng2015scalable}, PersonX \cite{sun2019dissecting}, and DukeMTMC-ReID \cite{ristani2016performance}, consistently demonstrate significant performance improvement compared to the baseline approach. The results highlight the effectiveness of pseudo label refinement in UDA-based object Re-ID.

\section{Related Work}
\label{sec:related_work}
Clustering-based pseudo label generation has gained significant attention in the field of self-supervised learning, particularly in the context of unsupervised object re-identification (Re-ID) \cite{ge2020mutual, yu2019unsupervised,chen2020self, wang2020unsupervised,zheng2022soft}. These approaches leverage clustering algorithms to assign pseudo labels to unlabeled data, which then serve as supervision for training deep neural networks.

To overcome the limitations of limited labeled data, self-supervised learning-based systems have emerged as a powerful solution \cite{jaiswal2020survey}. Recent advancements in self-supervised learning have introduced confidence-based pseudo-labeling techniques \cite{lee2013pseudo, arazo2020pseudo, li2021comatch, wang2022debiased}, where unlabeled data is used as targets for training models by repeatedly generating pseudo-labels. In our proposed approach, we go beyond the conventional pseudo-label generation by refining the generated clusters. We employ pseudo-labels from two consecutive epochs to generate refined soft and hard labels at each iteration, enhancing the accuracy of pseudo-labels in the self-supervised learning process.

One notable approach is the \textbf{Bottom-Up Clustering} (BUC) method proposed by Lin et al. \cite{lin2019bottom}. BUC utilizes iterative clustering to group feature vectors into clusters and assigns pseudo labels based on the cluster assignments. These pseudo labels are subsequently used to supervise the training of deep neural networks for Re-ID. Zeng et al. \cite{zeng2020hierarchical} introduced \textbf{Hierarchical Clustering with Local Distance} (HCLD) method that builds upon clustering-based pseudo labels. HCLD improves the clustering process by considering local distances between samples, resulting in more accurate cluster assignments. The refined pseudo labels obtained from HCLD are then utilized to guide the learning process in unsupervised Re-ID. 

Cheng et al. \cite{CHENG2022104493} contributed by proposing the Hierarchical Invariance Transfer (HIT) method, improving algorithm performance by selectively considering positive samples and introducing graded sample treatment. Additionally, the study introduces the Margin-Maximization Loss (MML) function to optimize HIT and reduce the impact of inaccurately identified challenging samples. These innovations collectively advance unsupervised person re-identification, providing novel solutions for refining pseudo labels and addressing challenges in sample treatment and identification accuracy. Zhao et al.\cite{ZHAO2023104786} proposed DHCL, a Dynamic Hybrid Contrastive Learning method for unsupervised person re-ID. It efficiently partitions the training dataset, guides feature extraction for intra-category similarity, and integrates contrastive learning levels to enhance feature space separability. A penalty item in the hybrid contrastive loss mitigates over-focusing on positive samples.

Another approach to tackle the problem of pseudo-label noise in person re-identification using Unsupervised Domain Adaptation (UDA) is \textbf{Learning with Noisy Labels} (LNL) proposed by Zhu et al. \cite{zhu2021learning}. It focuses on using Pseudo Label Correction (PLC), which employs sample similarity for fixing noisy pseudo labels. Furthermore, they provide a method for identifying noisy labels in deep learning called Noise Recognition, which is based on \textit{Similarity and Confidence Relationships} (SACR). By utilizing a re-weighting (RW) technique, they improved the training process with an Easy-to-Hard Model Collaborative Training (MCT) strategy that withstands noise and avoids overfitting. The objective of that method was to develop a more reliable training model for UDA-based person re-identification.

To further enhance the quality of pseudo labels, researchers have explored some advanced techniques \cite{yang2016joint,song2020learning,zhao2022exploiting}. For instance, Song et al. proposed the \textbf{Active Pseudo Labeling} (APL) method \cite{song2020learning}, which dynamically adjusts the pseudo label threshold during training. By adapting the threshold based on the confidence of pseudo labels, APL mitigates the negative impact of noisy pseudo labels, leading to improved learning in unsupervised Re-ID. Yang et al. introduced the \textbf{Joint Unsupervised Learning} (JUL) framework \cite{yang2016joint}, which integrates pseudo-label-based Re-ID with domain adaptation. JUL leverages pseudo labels from both the source and target domains to enhance the adaptation of learned features. By jointly optimizing Re-ID and domain adaptation objectives, JUL achieves impressive performance on challenging cross-domain Re-ID tasks.
During training, image clusters and representations are updated jointly:
image clustering is conducted in the forward pass, while representation learning is in the backward pass. The key idea behind their framework is that good representations are beneficial to image clustering and clustering results provide supervisory signals to representation learning.
Wang et al. \cite{wang2022unsupervised} employ a multi-branch network to extract local and global features and utilize a hierarchy-based clustering method for cluster generation.

\begin{figure*}[t]
  \begin{center}
  \includegraphics[trim=0cm 4cm 0cm 4cm, clip, width=1\textwidth]{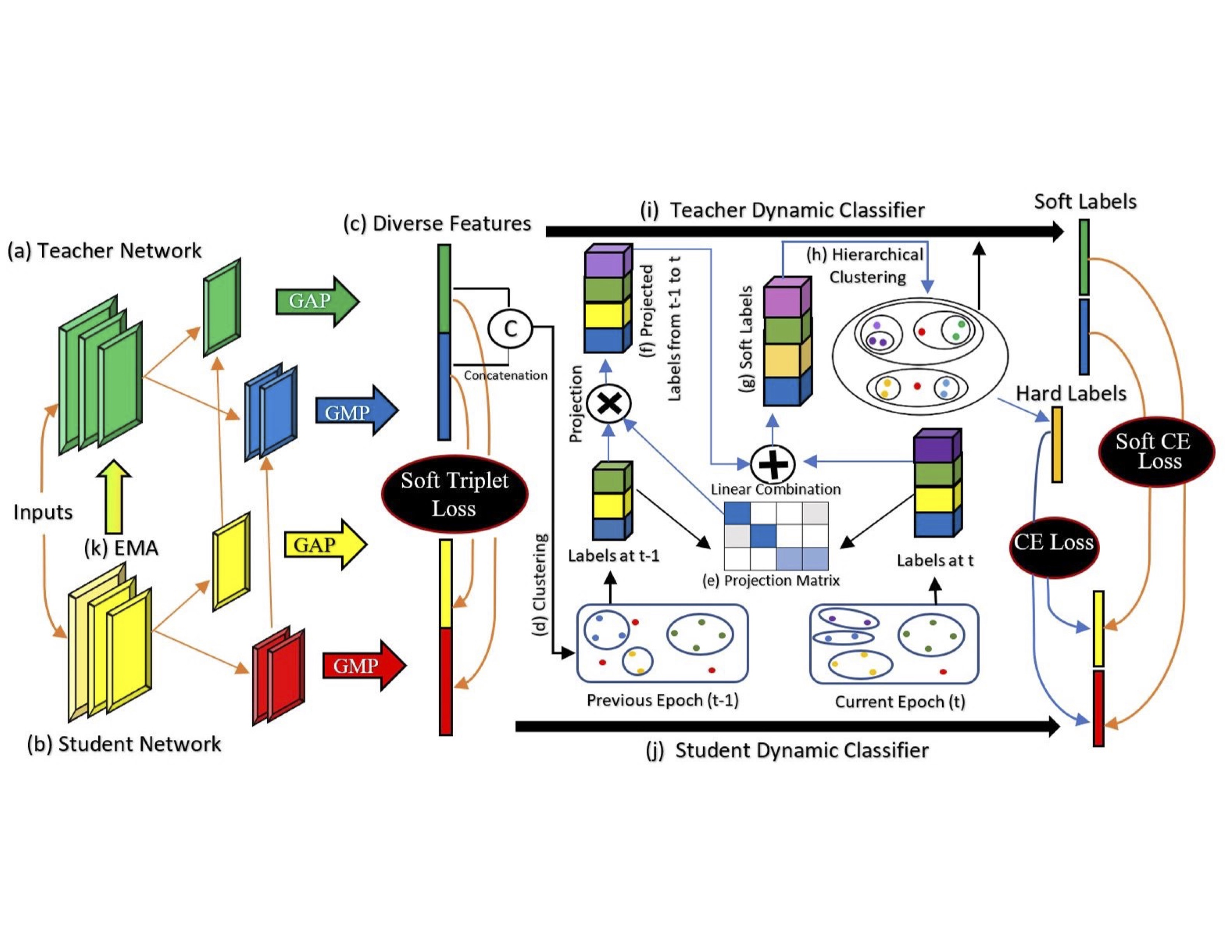}
  \end{center}
       \caption{System diagram of modified baseline with proposed Pseudo Label Refinement (SLR) algorithm: (a) Teacher network, (b) Student network, (c) Diverse feature layer, (d) The SLR algorithm starts here: clustering diverse features from teacher network, (e) Projection matrix estimation between consecutive epochs, (f) Projected pseudo-labels from previous epoch to the current epoch label space, (g) Refined soft labels for the current epoch, (h) The SLR algorithm ends here: hierarchical clustering for hard labels generation, (i) Teacher dynamic classifier trained on hard labels, (j) Student dynamic classifier trained on hard labels and soft labels from teacher network using cross-branch supervision. (k) The updated student network weights are used to update the teacher network weights using Exponential Moving Average (EMA) approach.}
      \label{fig:overall-pipline}
\end{figure*}

These existing works collectively demonstrate the effectiveness of clustering-based pseudo labels as supervision in self-supervised learning. By incorporating clustering algorithms and introducing various refinements, these approaches have made significant progress in unsupervised Re-ID, enhancing feature representations and narrowing the performance gap with supervised methods. Continued research in this direction holds immense potential for advancing self-supervised learning and addressing the challenges associated with unsupervised visual recognition tasks.

The primary challenge that clustering-based algorithms have addressed is the improvement of pseudo-label accuracy and the mitigation of noisy pseudo-labels. Our work aligns with this focus and enhances the baseline performance of unsupervised domain adaptation for object Re-ID. We introduce a technique that leverages a projection matrix to project cluster labels from the previous epoch to the current epoch label space, generating refined soft pseudo-labels that cover the entire range of class probabilities across all clusters. 
To handle the reduced dimensionality of refined soft pseudo-labels, we employ hierarchical DBSCAN, which produces superior results without incurring much computational overhead. In a wide range of experiments, the proposed algorithm has demonstrated excellent performance compared to the existing state-of-the-art methods. 


\section{Proposed Pseudo Label Refinement (SLR) Algorithm}
\label{sec: SLR}
The proposed SLR algorithm is integrated into an existing UDA baseline to evaluate its effectiveness in pseudo label noise cleaning.  The  UDA baseline learning system consists of two steps. In the first step, the backbone network is pre-trained on the source domain dataset. In the second step, target domain adaptation is performed.   The proposed SLR algorithm is integrated into the second stage, in modules (d)-(h) as shown in Fig. \ref{fig:overall-pipline}. 

\subsection{Teacher Network}
During pre-training, a backbone network consisting of four stages of ResNet50 with duplication of the fourth stage is used \cite{chen2021enhancing}. To make these duplicate stages different from each other, a bottleneck \cite{he2016deep} is added to one stage while the other is kept the same. After the bottleneck stage, Global Average Pooling (GAP) is applied, while in the other stage, Max Average Pooling (MAP) is used. Such an arrangement helps to extract diversified features. More pre-training details may be found in \cite{chen2021enhancing}.
We divided the modified baseline into different modules and discussed each in the following subsections.

As a result of source domain pre-training, a trained backbone network is obtained and its weights are loaded to initialize the teacher network,  as shown in Fig. \ref{fig:overall-pipline}(a).

\subsection{Student Network}
The same pre-trained backbone network discussed in the last section is used to initialize the student network. The student network is shown in Fig. \ref{fig:overall-pipline}(b).

\subsection{Diverse Features}
The target domain data is input to both the teacher and the student networks to obtain the diverse features (Fig. \ref{fig:overall-pipline}(c)). Soft Triplet Loss (STL) (Eq. \ref{eq:stri_am_10}) is used to train the diverse features layer of the student network. Only The diverse features from the teacher network are concatenated and input to the next module.

\subsection{Clustering Module}
The concatenated diverse features from the teacher network are input to the clustering module (Fig. \ref{fig:overall-pipline}(d)), which is DBSCAN in our case. Clustering is obtained on all training data in each epoch. The clustering module may generate a different number of clusters in consecutive epochs.  Clusters from the previous $t-1$ epoch, are used to refine labels for the current epoch, $t$, therefore label refinement starts in the second epoch.

\subsection{Projection Matrix Computation}
\vspace{-2mm}
A projection matrix is used to project the cluster labels in the previous epoch to the current epoch cluster label space. Both spaces may have different dimensionality due to different numbers of clusters. Therefore, the projection matrix must handle the difference in the number of clusters in both epochs as well as all-pairs inter-cluster similarity across the two epochs. This matrix acts as a mapping between the two cluster label spaces while ensuring clustering consistency. 
Let $m^t$ and $m^{t-1}$ be the number of clusters and $\Gamma^t$ and $\Gamma^{t-1}$ be the pseudo-labels at the previous and current epochs in the one-hot encoding scheme. Let $\Gamma^t_a$ and $\Gamma^{t-1}_b$ be the two distinct clusters in each list of pseudo-labels.  Intersection over union (IoU) between each such pair of clusters is used to compute the corresponding value in the projection matrix $P(a,b)$ as shown in Fig. \ref{fig:overall-pipline}(e):
\begin{equation}
  P(a,b) =  \frac{|\Gamma^{t-1}_a \cap \Gamma^{t} _b|} {|\Gamma^{t-1} _a  \cup \Gamma^{t} _b|}, 
  \label{eq:sim_matrix_2}
\end{equation}
where $|.|$ is the number of samples in a set. The projection  matrix is normalized as follows:
\begin{equation}
  \widehat{P}(a,b) =  \frac{P(a,b)} {\sum_{b=1}^{m^{t}} P(a,b)}.
  \label{eq:norm_sim_matrix_3}
\end{equation}
The normalized projection matrix $\widehat{P}(a,b) \in \mathcal{R}^{m_{t-1}\times m_t}$ is used as a mapping between the two cluster label spaces.
\subsection{Pseudo-labels Projection Module }
\vspace{-2mm}
Pseudo-labels obtained in the previous epoch are projected to the current labels space using the normalized projection matrix (Fig. \ref{fig:overall-pipline}(f)). 

Let $p$ be a sample from the previous stage, which has one hot encoded label ${y}^{t-1}_{p} \in \mathcal{R}^{m_{t-1}}$ obtained by clustering. It is transformed into the current epoch clustering space as follows: 
\begin{equation} 
  \widehat{y}_p^{t-1} =  \widehat{P}^\top y_p^{t-1} \in \mathcal{R}^{m^{t}}
  \label{eq:enhan_labels_4}
\end{equation}
The above generated transformed label $\hat{y}_p^{t}$ has the dimensions equal to labels generated at the current stage $m^t$.

\subsection{Soft Labels Refinement Module}
\vspace{-2mm}
The projected soft labels $\widehat{y}_p^{(t-1)}$ are used to refine the one-hot encoded current pseudo-labels ${y}_p^{t}$ via a linear combination of the two vectors:
\begin{equation} 
  \widetilde{y}_p^{t} =  \alpha  y_p^{t} + (1 - \alpha) \widehat{y}_p^{t-1} ,
  \label{eq:soft_labels_5}
\end{equation}
where $1 \ge \alpha \ge 0$ controls the amount of refinement. A higher value of alpha close to one assigns more weight to the current label and a smaller weight to the projected label. To show the effect of alpha value, We performed some experiments and added our results to the ablation study. The refined soft labels $\widetilde{y}_p^{t}$ (Fig. \ref{fig:overall-pipline}(g)) smooth the variations of label generation between two consecutive epochs.

Soft refined pseudo-labels can be used directly for supervising the student network and it also improved its performance compared to the existing baseline. We observe that its performance can be further improved by applying hierarchical clustering.

\subsection{Generation of Refined Hard labels}
To get hard labels (Fig. \ref{fig:overall-pipline}(h)), we propose hierarchical clustering instead of just thresholding the soft labels. Clustering at this step improves performance due to better utilization of the information contained in the refined soft labels. It also gives an opportunity to adjust the number of clusters by varying minimum cluster size.
We input $\widetilde{y}_p^{t}$ to a clustering algorithm to get refined hard pseudo-labels ${h}_p^t$. In this step, we employed hierarchical DBSCAN \cite{campello2013density} which produces better clusters compared to the DBSCAN algorithm. HDBSCAN does not require the user to input the $\epsilon$ parameter but rather searches for the best values depending on the data. Note that in the previous clustering step, flat DBSCAN was used due to its low computational cost over high-dimensional data. Since the refined soft pseudo-labels have reduced dimensionality compared to the diverse features, HDBSCAN produces better results without incurring computational overhead. The number of generated hard clusters may vary from epoch to epoch, resulting in a varying number of classes. To handle this issue, dynamic classifiers are trained, which adjust their size depending on the number of hard labels in each epoch.

\subsection{Teacher Dynamic Classifier}

The refined hard labels ${h}_p^t$ are used to supervise the teacher training process in the target domain using cross-entropy loss. (Fig. \ref{fig:overall-pipline}(i))   
\begin{equation} 
  L_{c} =  -\sum (h_p^t)^\top \log p_m(x_p^t) -\sum (h_p^t)^\top \log p_a(x_p^t), 
  \label{eq:training_label_7}
\end{equation}
where $p_m(\cdot)$ and $p_a(\cdot)$ are the soft labels generated by the teacher network for input $x_p^t$. The two teacher branches are independently updated using hard-label supervision.

\begin{table*}
  
    \caption{ Evaluation of the proposed SLR algorithm with the baseline ABMT and other existing  UDA person Re-ID methods for DukeMTMC-ReID to Market1501  and Market1501 to DukeMTMC-ReID. The reported performance of ABMT and SLR is obtained in our implementation with batch size 16 on a single GPU. The highest values are highlighted in red, while the second-highest values are indicated in blue.
  } 
  \label{tab:Table_D_M}
  
  \begin{tabular*}{1\textwidth}{@{\extracolsep{\fill}} l l l  c c c c  c c c c } 
  \toprule
  \multirow{2}{*}{UDA Methods} & \multirow{2}{*}{Backbone} & \multirow{2}{*}{References} & \multicolumn{4}{c}{Duke-to-Market} & \multicolumn{4}{c}{Market-to-Duke} \\
  \cmidrule(rl){4-7} \cmidrule(rl){8-11}
    
    \multicolumn{3}{c}{}  & mAP  & Rank1 & Rank5 & Rank10 & mAP & Rank1 & Rank5 & Rank10 \\   \midrule

    TAL-MIRN \cite{9495801} & R50 & TCSVT-22 & 39.95 & 73.08 & 86.34 & - & 41.34 & 63.53 & 76.62 & -\\
    
    ECN \cite{zhong2019invariance} & R50 & CVPR-19 & 43.0 & 75.1 & 87.6 & 91.6 & 40.4 & 63.3 & 75.8 & 80.4\\

    PDA-Net \cite{li2019cross} & R50 & ICCV-19 & 47.6 & 75.2 &  86.3 & 90.2  & 45.1 & 63.2 & 77.0 & 82.5\\
    
    PCB-PAST \cite{zhang2019self} & R50 & ICCV-19 & 54.6 & 78.4 & - & - & 54.3 & 72.4 & - & -  \\
    
    SSG \cite{fu2019self} & R50 & ICCV-19  & 58.3 & 80.0 & 90.0 & 92.4 & 53.4 & 73.0 & 80.6 & 83.2  \\

    MMCL \cite{wang2020unsupervised} & R50 & CVPR-20 & 60.4 & 84.4 & 92.8 & 95.0 & 51.4 & 72.4 & 82.9 & 85.0  \\

    ACT \cite{yang2020asymmetric} & R50 & AAAI-20 & 60.6 & 80.5 & - & - & 54.5 & 72.4 & - & -  \\

    ECN-GPP \cite{zhong2020learning} & R50 & TPAMI-20 & 63.8 & 84.1 & 92.8 & 95.4 & 54.4 & 74.0 & 83.7 & 87.4  \\

    JVTC+ \cite{li2020joint} & R50 & ECCV-20 & 67.2 & 86.8 & 95.2 & 97.1 & 66.5 & 80.4 & 89.9 & 92.2 \\

    AD-Cluster  \cite{zhai2020ad} & R50 & CVPR-20 & 68.3 & 86.7 & 94.4 & 96.5 & 54.1 & 72.6 & 82.5 & 85.5 \\

    HCN \cite{9410298} & R50 & TCSVT-22 & 70.2 & 90.2 & - & - & 57.3 & 78.9 & - & - \\

    MMT \cite{ge2020mutual} & R50 & ICLR-20 & 71.2 & 87.7 & 94.9 & 96.9 & 65.1 & 78.0 & 88.8 & 92.5 \\
    
    SDA \cite{9777862} & R50 & TNNLS-22 & 74.3 & 89.7 & 95.9 & 97.4 & 66.7 & 79.9 & 89.1 & 92.7  \\ 

    LNL \cite{zhu2021learning} & IBN-R50 & Neurocomp-21 & 75.2 & 88.9 & 95.7 & 97.3 & 62.5 & 77.4 & 88.1 & 90.6\\

    ABMT* \cite{chen2021enhancing} & R50 & WACV-21 & 76.3 & 91.9 & 96.7 & 98.1 & 63.2 & 78.0 & 88.2 & 91.2\\

    CD Mixup \cite{9756889} & R50 & TPAMI-23 & 77.1 & 90.7 & 96.3 & 97.7 & 69.0 & 82.3 & 90.8 & 93.2  \\
    
    Dual-Ref \cite{dai2021dual} & R50 & TIP-21 & 78.0 & 90.9 & 96.4 & 97.7 & 67.7 & 82.1 & 90.1 & 92.5 \\

    UNRN \cite{zheng2021exploiting} & R50 & AAAI-21 & 78.1 & 91.9 & 96.1 & 97.8 & 69.1 & 82.0 & 90.7 & \textcolor{blue}{93.5}  \\

    CD Mixup \cite{9756889} & IBN R50 & TPAMI-23 & 79.0 & 91.7 & 96.2 & - & \textcolor{blue}{70.1} & 83.2 & \textcolor{blue}{90.9} & -  \\

    GLT \cite{zheng2021group} & R50 & CVPR-21 & 79.5 & 92.2 & 96.5 & 97.8 & 69.2 & 82.0 & 90.2 & 92.8 \\

    ABMT*\cite{chen2021enhancing} & IBN-R50 & WACV-21 & 80.3 & 92.9 & 97.4  & 98.6 & 67.8 & 81.5 & 89.9 & 93.0 \\

    AWB \cite{9677903} & R50 & TIP-22 & 81.0 & \textcolor{blue}{93.5} & \textcolor{blue}{97.4} & \textcolor{blue}{98.3} & \textcolor{red}{70.9} & \textcolor{red}{83.8} & \textcolor{red}{92.3} & \textcolor{red}{94.0} \\

    PDGCN \cite{zhang2023local} & R50 & TOMM-23 & \textcolor{blue}{81.3} & 91.5 & 97.1 & 98.0 & 69.8 & \textcolor{blue}{83.6} & 90.1 & 92.3 \\
    
    \midrule
     
    \textbf{SLR (Ours)} & R50 &  &  {\textbf{78.9}} &  {\textbf{92.0}} &  {\textbf{96.9 }} &  {\textbf{98.0}} &  {\textbf{65.6}} &  {\textbf{ 79.4}} &  {\textbf{89.6}} &  {\textbf{92.7}} \\

    \textbf{SLR (Ours)} & IBN-50 & &  \textcolor{red}{\textbf{82.4}} & \textcolor{red}{\textbf{93.7}} & \textcolor{red}{\textbf{97.7}} & \textcolor{red}{\textbf{98.6}} & {\textbf{69.0}} & {\textbf{82.7}}  & {\textbf{90.5}} & {\textbf{93.1}} \\
    \bottomrule
  \end{tabular*}
  
\end{table*}

\subsection{Student Dynamic Classifier} 
Student network is trained using cross-branch supervision, as shown in Fig. \ref{fig:overall-pipline}j. It means each teacher branch is used to supervise a different student branch.
The student dynamic classifier is trained using hard labels $h_p^t$ with cross-entropy loss similar to Eq. \ref{eq:training_label_7} and student predictions $s_m(\cdot)$ and $s_a(\cdot)$ with soft labels generated by the teacher using cross-entropy loss, as shown below:
 $$ L_{sc} =  -\sum_{p} p_a(x_p^t)^\top \log s_m(x_p^t) + p_m(x_p^t)^\top \log s_a(x_p^t). 
  \label{eq:sce_am_8}$$
In addition to these two losses, Soft Triplet Loss (STL) \cite{dubourvieux2021unsupervised} is also used to train the diverse features layer of the student network, as shown in Fig. \ref{fig:overall-pipline}(b). The STL aims to minimize the distance between the diverse features of the same class and maximize it for different classes. It also helps the student network to distill knowledge from the teacher network.
 $$ L_{st} =  -\sum_{p} d_a^T(x_p^t)\log d_m^S(x_p^t)+ d_m^T(x_p^t)\log d_a^S(x_p^t), 
  \label{eq:stri_am_10}$$
where $d_a^T(\cdot)$ and $d_m^T(\cdot)$ are the soft triplet distances for the teacher network for GAP and GMP branches. In general, it is given by:
 $$   d(x_p^t) = \frac{\exp|| f(x_p^t) - f(x_{+}^t) ||_2} {\exp|| f(x_p^t) - f(x_{+}^t) ||_2 + \exp|| f(x_p^t|) - f(x_{-}^t) ||_2}.$$
In a mini-batch, the positive samples $x_{+}^t$ and negative samples $x_{-}^t$ are decided using the hard labels $h_p^t$. 

\subsection{Updating Teacher Weights} 
Once the student network is trained, its weights are used to update the teacher network weights using the exponential moving average (EMA), as shown in Fig. \ref{fig:overall-pipline}(k).

Overall loss for the target domain UDA training is given by:
\begin{equation} 
    \lambda_{c}^{t} L_{c}+\lambda_{sc}^{t} L_{sc}
     + \lambda_{st}^{t} L_{st}.
  \label{eq:overall_target_loss}
\end{equation}
The values of hyper-parameters are to be determined empirically.

\begin{table*} [t]
\centering
  \caption{Performance comparison of the proposed SLR algorithm with the existing state-of-the-art UDA person Re-ID methods on PersonX to Market1501  and Market1501 to PersonX.  The reported performance of ABMT is obtained by our implementation using a batch size of 16 on a single GPU (GeForce RTX 2080 Ti).  }
  \label{tab:Table_PX_M}

  \begin{tabular*}{1\textwidth}{@{\extracolsep{\fill}} l l l  c c c c  c c c c }  \toprule
  UDA Methods & Backbone & Reference  & \multicolumn{4}{c}{Market-to-PersonX} & \multicolumn{4}{c}{PersonX-to-Market} \\ \cmidrule(rl){4-7} \cmidrule(rl){8-11}
    
    \multicolumn{3}{c}{} & mAP  & Rank1 & Rank5 & Rank10 & mAP & Rank1 & Rank5 & Rank10 \\  \midrule

    ABMT* \cite{chen2021enhancing} & R50 & WACV-21 & 67.8 & 88.7 & 96.4  & 98.4 & 71.0 & 89.0 & 95.0 & 96.9 \\

    SPCL \cite{ge2020self} & R50& NeurIPS-20  & - & - & - & -  & 73.8 & 88.0 & 95.3 & 96.9 \\
     
    CD Mixup \cite{9756889} & R50 & TPAMI-23  & - & - & - & -  & 73.9 & 89.4 & 95.2 & 96.6 \\

    PDGCN \cite{zhang2023local} & R50 & TOMM-23 & - & - & - & -  & 75.7 & 87.7 & 95.3 & 97.1 \\

    ABMT* \cite{chen2021enhancing} & IBN-R50 & WACV-21 & 74.9 & 87.1 & 95.7  & 98.0 & \textcolor{blue}{\textbf{76.0}} & \textcolor{blue}{\textbf{90.9}} & \textcolor{blue}{\textbf{96.3}} & \textcolor{blue}{\textbf{97.7}} \\

    \midrule
    
    \textbf{SLR} (Ours) & R50 &  & \textcolor{blue}{\textbf{79.1}} & \textcolor{blue}{\textbf{92.6}} & \textcolor{blue}{\textbf{97.7}} & \textcolor{blue}{\textbf{98.9}} &  {\textbf{73.1}} &  {\textbf{89.0}}  &  {\textbf{95.2}} &  {\textbf{97.0}} \\

     \textbf{SLR} (Ours) & IBN-R50 & & \textcolor{red}{\textbf{86.1}} & \textcolor{red}{\textbf{94.2}} & \textcolor{red}{\textbf{98.4}} & \textcolor{red}{\textbf{99.3}} & \textcolor{red}{\textbf{78.9}} & \textcolor{red}{\textbf{91.7}}  & \textcolor{red}{\textbf{97.0}} & \textcolor{red}{\textbf{98.2}} \\

    \bottomrule
    
  \end{tabular*}
\end{table*}

\section{Experiments and Results}

\textbf{Datasets:}
Experiments are performed on three datasets including
Market1501 \cite{zheng2015scalable} consisting of 32668 images of 1501 identities. For training, 12936 images of 751 identities while for testing 19732 images of 750 identities are used. The second dataset DukeMTMC-ReID \cite{ristani2016performance}  consists of 36411 labelled images of 1404 identities where 16522 images of 702 identities are used for training and 19889 images of 702 identities are used for testing. The third is a synthetic dataset, PersonX \cite{sun2019dissecting}, consisting of 45,792 images featuring 1,266 distinct identities, where 9840 images of 410  identities are used as training, 30816 images of 856 identities are used as gallery data, and the remaining 5136 images of 856 identities are used for evaluation. 

\textbf{Evaluation Metrics:} Cumulative Matching Characteristics(CMC) \cite{gray2007evaluating} and mean Average Precision (mAP) \cite{zheng2015scalable} are used to compare results with existing state-of-the-art methods.

\textbf{Experimental Settings:}
Experiments are performed using two backbone models including ResNet50 and IBN-ResNet50 \cite{pan2018two}. The training is performed on a single GPU (GeForce RTX 2080 Ti) using Pytorch. Both backbones are pre-trained on the source dataset for 80 epochs, using a batch size of 16. The input images are resized to 256$\times$128 pixels. The learning rate is 0.00035 and multiplied by 0.10 at milestone epochs of 40 and 70.

For target domain adaptation, the backbones are trained for  50 epochs, learning rate is 0.00035 with a weight decay rate of 0.0005. For hierarchical clustering, we adapted HDBSCAN \cite{campello2013density} in which min-cluster-size is set to 7. In Equation \ref{eq:overall_target_loss}, the values of $\lambda_c^t$ and $\lambda_{sc}^t$ are set to 0.50, while $\lambda_{st}^t$ is set to 1.00, $\alpha = 0.90$ is used to get refined soft labels in Eq. \ref{eq:soft_labels_5}, the batch size is 16. Random erasing data augmentation is used similarly to \cite{zhong2020random}. The code of SLR can be found in supplementary material.

\subsection{Comparisons and Discussion}
We compare our proposed framework with baseline ABMT \cite{chen2021enhancing} and twenty existing state-of-the-art  UDA methods on the three person-Re-ID datasets. Experiments are conducted in four cross-settings: source domain to target domain. For fair comparisons, experiments are performed on two backbones including Resnet50 and IBN-Resnet50. The reported performance of ABMT is obtained in our implementation using a batch size of 16 due to the availability of only one GPU. A larger batch size on more GPUs may have resulted in improved performance of both ABMT and SLR.    \textbf{Table \ref{tab:Table_D_M}} shows the comparisons on DukeMTMC-to-Market1501 and Market1501-to-DukeMTMC. In the first setting, the modified baseline (SLR) has obtained an improvement of 2.6\% and 2.1\% mAP for both backbones. In the  Market1501-to-DukeMTMC experiment, SLR has obtained an improvement of 2.4\% and 1.2\% mAP respectively. In both experiments, SLR has also obtained better performance over most of the compared  UDA methods. Some compared methods such as AWB and MMT used a batch size of 64 while CD Mixup used a batch size of 256 resulting in performance improvement. 
\textbf{Table \ref{tab:Table_PX_M}} shows the comparisons on Market1501 to PersonX and PersonX to Market1501. In the first setting, SLR has obtained a significant improvement of 11.3\% and 11.2\% respectively. In the second setting, SLR has obtained an improvement of 2.1\% and 2.9\% respectively. In these experiments, SLR has obtained improved performance than the four existing methods.
The proposed SLR algorithm has consistently improved person Re-ID performance in UDA in the four experiments over three datasets and using mAP, Rank 1, 5 \& 10 accuracy. These results demonstrate the effectiveness of the SLR algorithm in improving the pseudo labels across different domains.

\subsection{Ablation Studies}

\subsubsection{Clustering Algorithms in Module 3.(d) \& 3.(h)}

Table \ref{tab:clustering_module_d}  compares the performance of the proposed SLR algorithm using different clustering options.  If DBSCAN is used in both modules, performance degrades by 1.9\% mAP. This is because DBSCAN has not performed well on significantly reduced dimensionality in Fig. 2(h). If HDBSAN is used in both modules, again a performance degradation of 1.3\% is observed.
This is because HDBSCAN was not able to effectively handle highly diverse feature dimensionality. Also, the computational complexity of HDBSCAN is significantly larger than DBSCAN on high-dimensional data. These ablation experiments justify the use of DBSCAN in module Fig. 2(d) and hierarchical DBSSCAN in module Fig. 2(h). 

\begin{table}[t]
\centering
  \caption{Performance of SLR on the Duke-to-Market setting by switching clustering algorithms in Module 3.(d) and 3.(h)}
  \label{tab:clustering_module_d}
  
  \begin{tabular}{c c  c c }  \toprule
  \multicolumn{2}{c}{Clustering}& \multicolumn{2}{c}{Duke-to-Market}  \\ \cmidrule(rl){1-2} \cmidrule(rl){3-4}
    
    Module 3.(d) & Module (h)&  mAP & Rank1 \\   \midrule

    DBSCAN & - & 80.4 & 93.0   \\
    DBSCAN & DBSCAN & 80.5 & 93.2   \\
    DBSCAN & MaxThresh & 80.9 & 93.1   \\
    HDBSCAN & HDBSCAN & \textcolor{blue}{81.1} & \textcolor{blue}{93.6}   \\
    DBSCAN & HDBSCAN & \textcolor{red}{\textbf{82.4}} & \textcolor{red}{\textbf{93.7}}  \\

    \bottomrule
  \end{tabular}
    
\end{table}

\subsubsection{Using Max Threshold in Module 3(h)}
In Module 3(h), instead of using HDBSCAN, if only maximum is used to get the hard labels, we get mAP of 80.9 and Rank 1 accuracy of 93.1 (see Table \ref{tab:clustering_module_d}). This experiment demonstrates the importance of using HDBSCAN in this module.

\subsubsection{Elimination of Module 3(h)}
Table \ref{tab:clustering_module_d} compares the performance of the proposed SLR algorithm with IBN-Resnet50 as the backbone by eliminating Module 3(h). As a result, the performance reduces by 2.2\% mAP. This experiment highlights the contribution of refined hard labels to the overall performance.

\begin{table} [b]
\centering
  \caption{Performance of the  SLR  on the Duke-to-Market setting by varying the minimum cluster size in HDBSCAN algorithm in module 3.(h).}
  \label{tab:min_cluster_size}
  \begin{tabular}{c   c  c  c  c}  \toprule
  \multirow{1}{*}{Minimum} & \multicolumn{4}{c}{Duke-to-Market}  \\ \cline{2-5}
    
   Cluster Size & mAP & Rank1 & Rank5 & Rank10 \\   \midrule
    
    4 & 80.5 & 93.2 & 97.7 & 98.3  \\
    5 & 81.0 & 93.3 & 97.5 & 98.4  \\
    6 & \textcolor{blue}{82.1} & \textcolor{blue}{93.4} & \textcolor{blue}{97.6} & \textcolor{blue}{98.5} \\
    7 & \textcolor{red}{\textbf{82.4}} & \textcolor{red}{\textbf{93.7}} & \textcolor{red}{\textbf{97.7}} & \textcolor{red}{\textbf{98.6}} \\
    8 & 81.9 & 93.0 & 93.2  & 93.2\\    \bottomrule
  \end{tabular}
\end{table}

\subsubsection{Effect of Minimum Cluster Size in Module 3(h)} 
The effect on the performance of the model by varying minimum cluster size parameters in HDBSCAN in Module 3(h) is also studied. As illustrated in Table \ref{tab:min_cluster_size}, min cluster size is varied as 4, 5, 6, 7, and 8, and performance is computed in Duke to Market settings. We observed the best performance for a cluster size of 7, therefore the same size is used to report all results.

The number of clusters generated by increasing min cluster size from 4 to 8 is found to be decreasing.  
For min-cluster=7, SLR obtained 670 clusters in Market as target (Duke to Market), 673 for (PersonX to Market),  828 clusters in Duke (Market to Duke), and 424 in PersonX (Market to PersonX). The number of obtained clusters is closer to the number of identities in each dataset, compared to the baseline approach. This fact also gives an important insight into the performance improvement by the proposed SLR algorithm.

\subsubsection{Ablation on Loss Terms} 
To evaluate the effect of each loss term on the performance of the SLR, we remove each loss term in Eq. \ref{eq:overall_target_loss} one by one and report the results in Table \ref{tab:wo_loss}. We note that if $L_c$ loss is removed, then the performance decreases by 10.4\% which is because $L_c$ directly uses the hard refined labels to supervise the student predictions. Removal of $L_{sc}$ resulted in 0.60\%, and removal of $L_{st}$ resulted in a performance reduction of 2.0\%. These experiments demonstrate each loss term's contribution to the final performance.

\begin{table}[t]
\centering
\caption{ Performance variation of the SLR algorithm by eliminating each loss term one by one. }
  \label{tab:wo_loss}
  
  \begin{tabular}{ c  c  c  c  c }  \toprule
  \multirow{2}{*}{Loss } & \multicolumn{4}{c}{Duke-to-Market}  \\ \cmidrule(rl){2-5}
    
    & mAP & Rank1 & Rank5 & Rank10 \\   \midrule
    SLR & \textcolor{red}{82.4} & \textcolor{red}{93.7} & \textcolor{red}{97.7} & \textcolor{red}{98.6}  \\
    SLR w/o $L_{c}$ & 72.0 & 87.4 & 93.8 & 95.9 \\
    SLR w/o $L_{sc}$ & \textcolor{blue}{81.8} & \textcolor{blue}{92.9} & \textcolor{blue}{97.3}  & 98.2 \\
    SLR w/o $L_{st}$ & 80.4 & 92.9 & 97.2  & \textcolor{blue}{98.4} \\
    \bottomrule
  \end{tabular}

\end{table}

\begin{table}[h]
\centering
\caption{ Performance variation of the SLR algorithm by varying the values of $\lambda$ in \ref{eq:overall_target_loss} }
  \label{tab:lambda_values}
  
  \begin{tabular}{ c  c  c  c c}  \toprule
  \multirow{2}{*}{ $\lambda$ Values } & \multicolumn{4}{c}{Duke-to-Market}  \\ \cmidrule(rl){2-5}
  
    & mAP & Rank1 & Rank5 & Rank10 \\   \midrule  
    
    $\lambda_{st}^t$ = $\lambda_c^t$ = $\lambda_{sc}^t = 1$& 80.3 & 93.0 & 97.7 & 98.5 \\
    
    $\lambda_{st}^t = \lambda_c^t$ = $\lambda_{sc}^t= 0.5$  & \textcolor{blue}{81.6} & \textcolor{blue}{93.6} & \textcolor{red}{97.8} & \textcolor{blue}{98.6}  \\
    
    $\lambda_c^t =\lambda_{sc}^t=0.50$ , $\lambda_{st}^t=1$ &  \textcolor{red}{82.4} & \textcolor{red}{93.7} & \textcolor{blue}{97.7} & \textcolor{red}{98.6} \\
    \bottomrule
  \end{tabular}

\end{table}

An ablation study is also performed on the relative importance of each loss term in Eq. \ref{eq:overall_target_loss}. Experiments are performed by varying the values of  $\lambda$ as shown in Table \ref{tab:lambda_values}. Best performance is obtained for $\lambda_c^t=\lambda_{sc}^t=0.50$, and $\lambda_{st}^t=1.0$. All results in this work are reported for these settings. These experiments demonstrate that further fine-tuning of these parameters may have yielded improved performance of the proposed SLR algorithm.

\subsubsection{Effect on Performance by Varying the Number of Training Epochs }
\begin{figure}[h]
  \centering
  \includegraphics[width=3.0in]{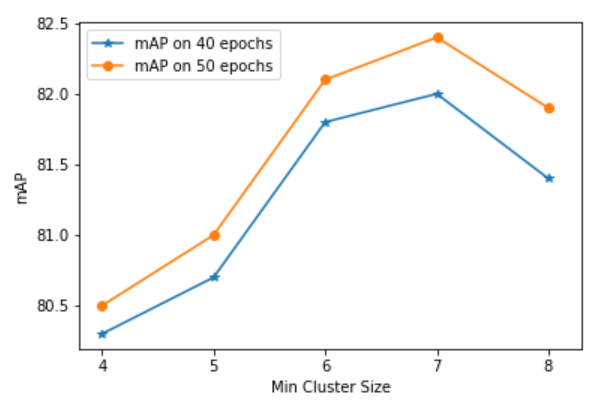}
  \caption{Performance comparison of SLR algorithm in terms of mAP on  40 and 50 epochs with varying min cluster size parameter. 
  }
  \label{fig:epochs_map}
\end{figure}

\begin{figure}[h]
  \centering
  \includegraphics[width=3.0in]{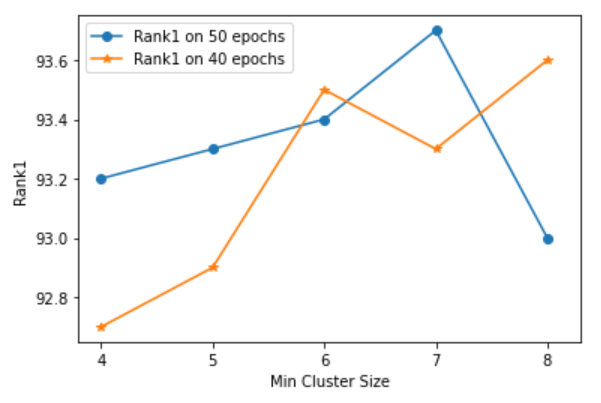}
  \caption{Performance comparison of SLR algorithm in terms of rank 1 accuracy on 40 and 50 epochs. The best result for rank 1 is 93.7\% on 50 epochs and a minimum cluster size of 7.  }
  \label{fig:epochs_rank}
\end{figure}

Fig. \ref{fig:epochs_map} compares mAP for the SLR algorithm over varying values of min-cluster size for 40 and 50 epochs. For all experiments, performance at 50 epochs is improved compared to 40 epochs. Fig. \ref{fig:epochs_rank} shows the corresponding rank-1 performance. These experiments demonstrate improved performance for 50 epochs.  

\section{Conclusion}
In this work, we have proposed a pseudo-label refinement (SLR) algorithm to enhance the performance of self-learning systems. Our algorithm focuses on reducing pseudo-label noise generated by the clustering approaches and also due to the domain gap. By introducing soft pseudo-labels and utilizing a cluster-label projection matrix, we improve cluster consistency across consecutive training iterations. Additionally, we employ hierarchical clustering to generate hard-refined labels, enabling enhanced supervision during training. We have evaluated the effectiveness of the SLR algorithm in the context of person Re-ID using an existing baseline approach. The results demonstrate significant performance improvement compared to the existing state-of-the-art methods. The modified baseline, empowered by our SLR algorithm, outperforms other approaches in terms of accuracy and robustness.

It is important to highlight that our SLR algorithm is not limited to person Re-ID but can be applied to enhance any self-supervised learning system. By reducing pseudo-label noise and improving cluster consistency, the SLR algorithm has the potential to enhance the performance of various applications across different domains. Overall, our research contributes to the advancement of self-learning systems by introducing a generic and effective algorithm for pseudo-label refinement. We believe that our findings pave the way for future improvements in self-supervised learning and open up new possibilities for enhancing the performance of a wide range of applications.


\end{document}